\documentclass[a4paper, 10pt, conference]{ieeeconf}      
\usepackage{FG2026}

\FGfinalcopy 

\IEEEoverridecommandlockouts                              
\usepackage{graphics} 
\usepackage{amsmath} 
\usepackage{amssymb}  
\usepackage{booktabs}
\usepackage{multirow}
\usepackage{hyperref}
\usepackage{subcaption}
\usepackage{tikz}
\usepackage{import}

\usepackage{xspace}

\definecolor{myblue}{RGB}{0,102,204}
\definecolor{myred}{RGB}{220,20,60}

\def\FGPaperID{274} 

\title{\LARGE \bf
ATTN-FIQA: Interpretable Attention-based Face Image Quality Assessment with Vision Transformers
}

\author{\parbox{16cm}{\centering
    {\large Guray Ozgur$^{1,2}$, Tahar Chettaoui$^{1,2}$, Eduarda Caldeira$^{1,2}$, Jan Niklas Kolf$^{1,2}$, Marco Huber$^{1,2}$, Andrea Atzori$^{1}$, Naser Damer$^{1,2}$, Fadi Boutros$^{1}$}\\
    {\normalsize
    $^1$ Fraunhofer Institute for Computer Graphics Research IGD, Darmstadt, Germany\\
    $^2$ Department of Computer Science, TU Darmstadt, Germany}
    \thanks{This research work has been funded by the German Federal Ministry of Education and Research and the Hessen State Ministry for Higher Education, Research and the Arts within their joint support of the National Research Center for Applied Cybersecurity ATHENE.}}
}

\usepackage{xspace}
\usepackage[table,xcdraw]{xcolor}
\newcommand{\ourmethod}{ATTN-FIQA\xspace}

\begin{document}

\maketitle
\pagestyle{plain}

\begin{abstract}
Face Image Quality Assessment (FIQA) aims to assess the recognition utility of face samples and is essential for reliable face recognition (FR) systems. Existing approaches require computationally expensive procedures such as multiple forward passes, backpropagation, or additional training, and only recent work has focused on the use of Vision Transformers. Recent studies highlighted that these architectures inherently function as saliency learners with attention patterns naturally encoding spatial importance. This work proposes ATTN-FIQA, a novel training-free approach that investigates whether pre-softmax attention scores from pre-trained Vision Transformer-based face recognition models can serve as quality indicators. We hypothesize that attention magnitudes intrinsically encode quality: high-quality images with discriminative facial features enable strong query-key alignments producing focused, high-magnitude attention patterns, while degraded images generate diffuse, low-magnitude patterns. ATTN-FIQA extracts pre-softmax attention matrices from the final transformer block, aggregate multi-head attention information across all patches, and compute image-level quality scores through simple averaging, requiring only a single forward pass through pre-trained models without architectural modifications, backpropagation, or additional training. Through comprehensive evaluation across eight benchmark datasets and four FR models, this work demonstrates that attention-based quality scores effectively correlate with face image quality and provide spatial interpretability, revealing which facial regions contribute most to quality determination. The implementation is publicly available at: \begin{footnotesize}
    \url{https://github.com/gurayozgur/ATTN-FIQA} 
\end{footnotesize}\vspace{-2mm}
\end{abstract}

\thispagestyle{plain}

\begin{figure}[t]
    \centering
    \includegraphics[width=0.48\textwidth]{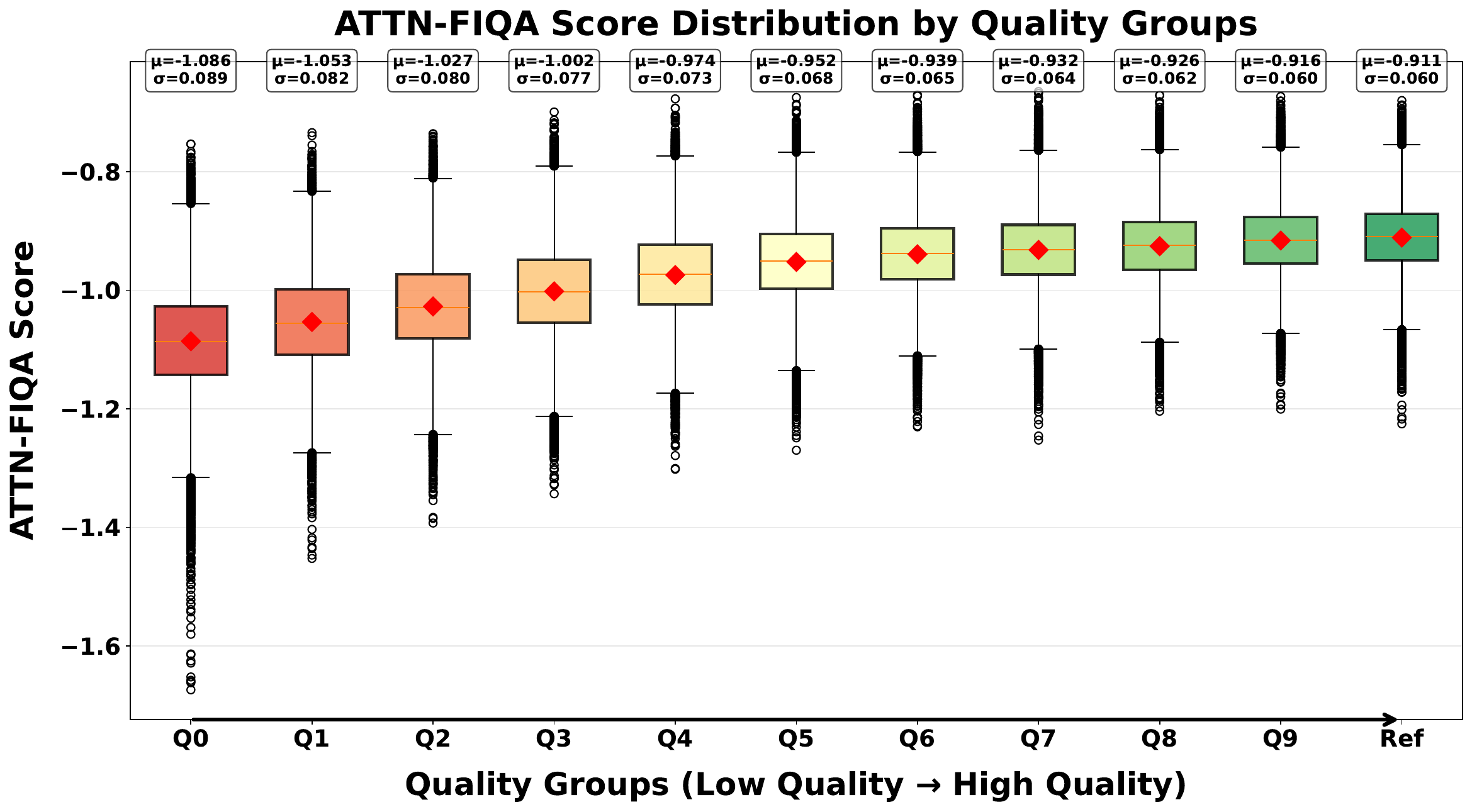}
    \caption{Empirical validation of \ourmethod on SynFIQA \cite{mrfiqa}. Distribution of \ourmethod scores across 11 quality groups (Q0-Q9: degraded images from lowest to highest quality; Ref: reference images) for 550,000 synthetic face images with ground-truth labels. Each boxplot shows the distribution of predicted scores for images within that group. The plot demonstrates clear discriminative ability: Q0 exhibits the lowest mean score, which monotonically increases through Q1-Q9, with Ref achieving the highest scores. This systematic correlation between predicted scores and ground-truth quality groups validates our hypothesis that pre-softmax attention magnitudes intrinsically encode face image quality.}
    \label{fig:synfiqa_validation} \vspace{-6mm}
\end{figure}

\begin{figure*}[ht]
    \centering
    \includegraphics[width=0.9\textwidth]{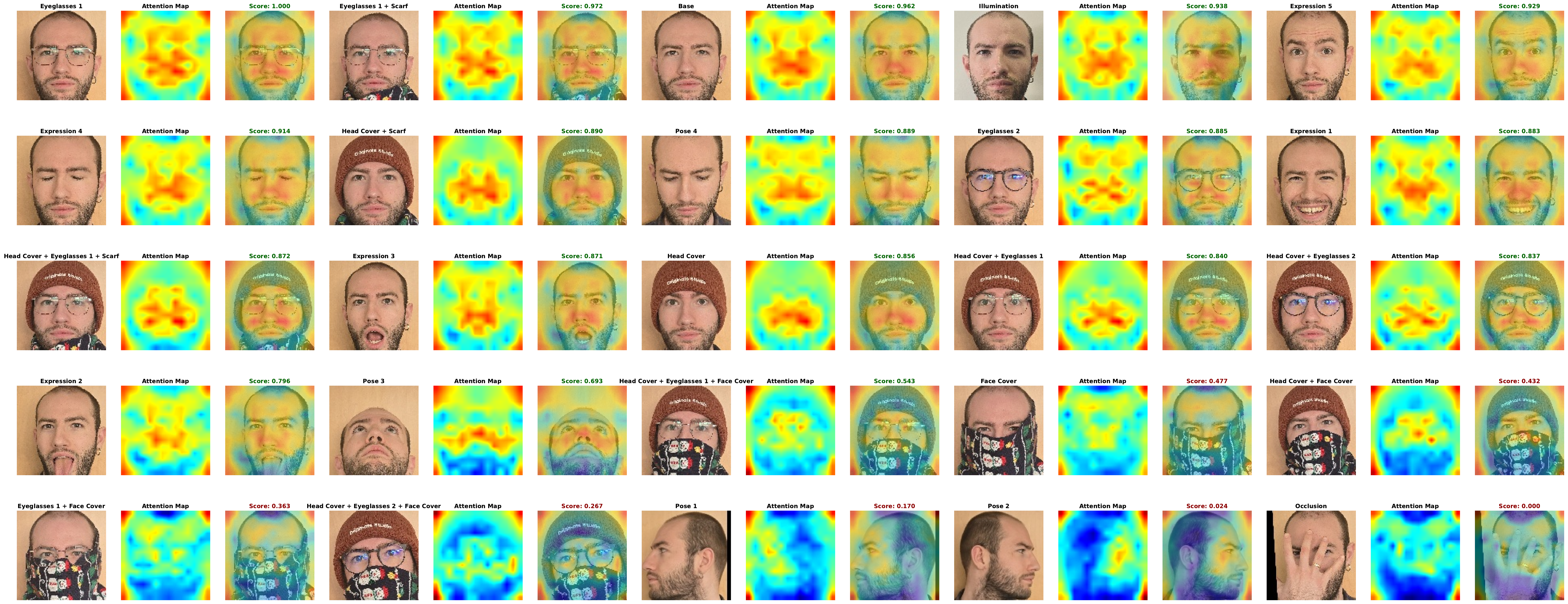}
    \caption{Attention visualization for controlled quality conditions using ViT-S/WebFace4M/AdaFace model. Each row contains 5 images, with each image displayed in three columns: (left) original image with condition label, (center) pre-softmax attention heatmap with global colormap, (right) overlay with normalized quality score. Images are sorted by quality score from top-left (highest) to bottom-right (lowest). Quality scores are normalized to [0,1] across all images in this figure. The global colormap (\textcolor{myblue}{blue}=low attention magnitude, \textcolor{myred}{red}=high attention magnitude) reveals that high-quality frontal poses produce focused, high-magnitude attention (\textcolor{myred}{red} regions) on discriminative facial features, while degraded conditions (occlusions, extreme poses, face covers) exhibit diffuse, low-magnitude attention (\textcolor{myblue}{blue} regions), reflecting model uncertainty.}
    \label{fig:controlled_vits_adaface} \vspace{-6mm}
\end{figure*}

\section{INTRODUCTION}

FIQA quantifies recognition utility, measuring how effectively a face image serves automated identity verification \cite{Quality_ISO,NISTQuaity,DBLP:journals/csur/SchlettRHGFB22}. Current SOTA methods fall into four three categories: (1) \textit{supervised approaches} that train quality regressors from explicit or proxy labels \cite{faceqnetv1, SDDFIQA, best2018learning, RANKIQ_FIQA}, (2) \textit{unsupervised approaches} that estimate quality from fixed models without FIQA-specific supervision, including diffusion-based and FR-analysis methods \cite{10449044, babnikTBIOM2024, SERFIQ, grafiqs}, and (3) \textit{self-supervised (FR-integrated) approaches} that couple quality estimation with FR model learning \cite{boutros_2023_crfiqa, MagFace, PFE_FIQA, atzori2025vitfiqaassessingfaceimage}. While these methods achieve strong performance, they typically require computationally expensive operations such as multiple forward passes, backpropagation, or additional training procedures. Furthermore, most approaches produce opaque scalar quality scores without providing spatial interpretability, limiting their utility for understanding which image regions contribute to quality degradation in practical biometric systems.

Vision Transformers (ViTs) have demonstrated strong performance in FR through attention-based architectures that model global patch relationships \cite{DBLP:conf/iclr/DosovitskiyB0WZ21,Dan_2023_TransFace,DBLP:conf/cvpr/KimS0JL24}. Recent work reveals that ViTs inherently function as saliency learners, with attention patterns naturally encoding spatial importance without task-specific training \cite{DBLP:conf/bmvc/DjilaliMO23}. Also, attention statistics have successfully served as utility proxies in applications such as out-of-distribution (OOD) detection and bias identification \cite{DBLP:journals/corr/abs-2210-14391, DBLP:conf/iccvw/CultreraSB23, DBLP:conf/cvpr/SerianniZRR25}. Despite this potential, existing ViT-based FIQA methods \cite{atzori2025vitfiqaassessingfaceimage} require additional training with learnable quality tokens and architectural modifications. This raises a fundamental question: \textit{Can pre-softmax attention scores from pre-trained ViT models directly serve as face image quality indicators without additional training?}

Drawing from these insights, we hypothesize that pre-softmax attention magnitudes intrinsically encode face image quality as a byproduct of the recognition task. High-quality images with discriminative facial features enable the model to compute strong query-key similarities, producing focused attention patterns with higher magnitudes across identity-relevant regions. Conversely, quality degradations such as blur, occlusion, or poor illumination introduce patch ambiguity that weakens query-key alignments, resulting in diffuse attention patterns with lower magnitudes. Unlike post-softmax normalization that constrains values to probability distributions, pre-softmax scores preserve these magnitude differences, making them particularly suitable for quality assessment. Extracting pre-softmax attention matrices from the final transformer block of pre-trained ViT-based FR models, we aggregate information across all attention heads and patches, and compute image-level quality scores through simple averaging. This approach requires only a single forward pass without architectural modifications, backpropagation, or additional training. We present \ourmethod (\textit{\textbf{Att}e\textbf{n}tion-based \textbf{FIQA}}), making two key contributions:
\begin{itemize}
    \item We demonstrate that pre-softmax attention scores from pre-trained ViT-based FR models effectively serve as quality indicators, requiring only a single forward pass.
    \item We provide spatial quality interpretability through attention-based assessment, revealing spatial indications of which facial regions contribute most to quality determination and enabling practitioners to understand the causes of quality degradation.
\end{itemize}


\section{RELATED WORK}
\label{sec:related}

\subsection{Attention Mechanisms and Interpretability}
\label{subsec:related_attention}

\textbf{Attention for Model Interpretability:} 
The self-attention mechanism computes content-dependent interactions between all pairs of tokens (image patches), producing per-head attention weight matrices over patch tokens. \cite{DBLP:conf/nips/VaswaniSPUJGKP17,DBLP:conf/iclr/DosovitskiyB0WZ21}. Various methods have been developed to interpret these attention patterns, including attention rollout, which propagates attention through the network by composing per-layer attention matrices across layers (accounting for residual connections) \cite{DBLP:conf/acl/AbnarZ20}, gradient-based attribution approaches such as Grad-CAM \cite{DBLP:journals/ijcv/SelvarajuCDVPB20}, relevance-propagation transformer attributions that propagate relevance through attention and feed-forward blocks \cite{DBLP:conf/cvpr/CheferGW21}, and causal analysis techniques that derive patch-level masks from patch-embedding features and estimate causal effects via image masking/perturbations rather than directly treating attention weights as explanations \cite{DBLP:conf/ijcai/Xie0CZ23}. These interpretability methods often treat attention-derived quantities as relevance proxies, and attribution quality depends strongly on how attention is aggregated across heads/layers and residual paths. Djilali et al. \cite{DBLP:conf/bmvc/DjilaliMO23} analyze raw CLS-to-patch attention maps from ViTs on saliency benchmarks and show that these maps can localize salient objects, with simple post-processing improving saliency quality.

\textbf{Attention as Task Utility Proxy:} Beyond post-hoc visualization, attention-weight distributions have been explored as quantitative signals related to prediction reliability in specific settings (e.g., detection/tracking). Ruppel et al. \cite{DBLP:journals/corr/abs-2210-14391} propose an attention spread metric derived from the covariance of the top-k cross-attention weights and analyze its relationship with detection accuracy. TAVAC \cite{doi:10.1126/sciadv.abg0264} evaluates overfitting by measuring how consistent attention-based explanations are for the same images when they appear in the training set versus the validation set, and uses lower consistency as an indicator of overfitting and less reproducible interpretations. Cultrera et al. \cite{DBLP:conf/iccvw/CultreraSB23} perform OOD detection by extracting ViT attention heatmaps and training an autoencoder to reconstruct them, using the reconstruction error as the OOD score, while Serianni et al. \cite{DBLP:conf/cvpr/SerianniZRR25} performed bias identification through Attention-IoU, which uses attention maps to quantify spurious correlations by comparing attention overlap with feature masks or with attention maps of correlated attributes, revealing confounding cues beyond accuracy disparities. In Image Quality Assessment (IQA), TRIQA \cite{11084443} proposes contrastive pretraining on ordered distortion triplets using content-aware and quality-aware ConvNeXt \cite{9879745} backbones to learn quality-sensitive representations. CQA-Face \cite{DBLP:conf/aaai/WangG22} introduces contrastive attention learning to encourage diverse attended facial parts and a quality-aware network that estimates part-level quality, improving robustness under occlusion, blur, illumination, and pose. These works collectively demonstrate that ViT attention patterns capture intrinsic properties of the input-model interaction, serving as valuable utility proxies for downstream task performance beyond their explanatory value. A critical distinction in attention-based approaches is the use of pre-softmax versus post-softmax attention scores. Pre-softmax attention, computed as the scaled dot-product of query and key matrices before applying the softmax function, preserves magnitude information about raw patch similarities that becomes unavailable after post-softmax normalization constrains values to probability distributions \cite{DBLP:conf/nips/VaswaniSPUJGKP17}. Unlike prior interpretability-focused approaches that primarily treat attention as a post-hoc explanation mechanism, our work investigates whether pre-softmax attention can directly serve as a quality signal for FIQA, leveraging the preserved magnitude information to quantify recognition utility.

\subsection{Face Image Quality Assessment}
\label{subsec:related_fiqa}

FIQA methods can be categorized into three paradigms: \textbf{(1) Supervised approaches} train quality estimators using explicit labels. FaceQnet~\cite{faceqnetv1} uses ICAO compliance standards as quality references, SDD-FIQA~\cite{SDDFIQA} employs distribution distances between embeddings, RankIQ~\cite{RANKIQ_FIQA} adopts a learning-to-rank strategy, training models to predict quality rankings based on FR performance metrics across different datasets, and CLIB-FIQA~\cite{Ou_2024_CVPR} incorporates additional supervision from CLIP \cite{DBLP:conf/icml/RadfordKHRGASAM21}. \textbf{(2) Unsupervised approaches} encompass both general and FR-specific methods. DifFIQA~\cite{10449044} leverages diffusion models to assess embedding robustness by exploring the stability of face representations under different noise conditions, while eDifFIQA~\cite{babnikTBIOM2024} distills this approach into a lighter model for faster inference. FR-specific methods operate on frozen FR models: SER-FIQ~\cite{SERFIQ} measures embedding stability under dropout perturbations by evaluating consistency with varied dropout patterns across multiple forward passes. GraFIQs~\cite{grafiqs} uses gradient magnitudes during backpropagation to evaluate how strongly each sample aligns with the FR model's optimization objective. FaceQAN~\cite{FaceQAN} estimates quality by quantifying adversarial robustness. Terhörst et al.~\cite{10088447} extend gradient-based approaches to pixel-level quality assessment for enhanced interpretability. ViTNT-FIQA~\cite{Ozgur_2026_WACV} tracks embedding consistency across ViT blocks. \textbf{(3) Self-supervised approaches} jointly optimize FR and FIQA objectives. MagFace~\cite{MagFace} links quality scores to embedding magnitudes through adaptive margin penalties and magnitude-aware losses. PFE~\cite{PFE_FIQA} models face embeddings as Gaussian distributions where the uncertainty represents quality. CR-FIQA~\cite{boutros_2023_crfiqa} estimates quality by predicting a sample's relative classifiability. ViT-FIQA~\cite{atzori2025vitfiqaassessingfaceimage} extends standard ViT backbones with a learnable quality token designed to predict utility scores. 

Despite these advances, existing approaches primarily exploit quality-relevant information from feature representations or require multiple forward passes, backpropagation, or separate training procedures. The interpretability of quality predictions remains limited, as most methods produce scalar quality scores without providing insights into which image regions or features contribute to the quality assessment. In this work, we investigate whether attention mechanisms within pre-trained ViT-based FR models can serve as quality indicators, requiring only a single forward pass while offering insights into the model's decision-making process through attention pattern analysis.


\section{METHODOLOGY}

ViTs process facial images through hierarchical attention mechanisms, with recent findings revealing that attention patterns naturally encode spatial importance and can serve as task utility proxies, as discussed in Section \ref{sec:related}. Building on these insights, this section presents our investigation of pre-softmax attention scores as quality indicators for FIQA. We start by establishing the mathematical foundations of ViT attention mechanisms, followed by our attention-based quality assessment framework.

\subsection{Preliminaries: Vision Transformer Architecture}
\label{subsec:vit_prelim}

Consider a ViT architecture \cite{DBLP:conf/iclr/DosovitskiyB0WZ21} for FR. Given an input face image $\mathbf{I} \in \mathbb{R}^{H \times W \times 3}$ (height $H$, width $W$, RGB channels), the image is divided into non-overlapping patches of size $P \times P$, resulting in $N = \frac{HW}{P^2}$ patches. Each patch is linearly projected to an embedding of dimension $D$:
\begin{equation}
    \mathbf{z}_0 = [\mathbf{Y}\mathbf{p}_1 + \mathbf{b}; \mathbf{Y}\mathbf{p}_2 + \mathbf{b}; \ldots; \mathbf{Y}\mathbf{p}_N + \mathbf{b}] + \mathbf{E}_{pos},
\end{equation}
where $\mathbf{Y} \in \mathbb{R}^{D \times (P^2 \cdot 3)}$ is the patch embedding projection matrix, $\mathbf{p}_i \in \mathbb{R}^{P^2 \cdot 3}$ is the $i$-th flattened patch, $\mathbf{b} \in \mathbb{R}^{D}$ is the bias term, and $\mathbf{E}_{pos} \in \mathbb{R}^{N \times D}$ are learnable positional embeddings. The embedded patches $\mathbf{z}_0 \in \mathbb{R}^{N \times D}$ are processed through $L$ transformer blocks. Each block $\ell \in \{1, \ldots, L\}$ applies multi-head self-attention (MSA) followed by a multi-layer perceptron (MLP) with residual connections:
\begin{equation}
\begin{aligned}
    \mathbf{z}'_\ell &= \text{MSA}(\text{LN}(\mathbf{z}_{\ell-1})) + \mathbf{z}_{\ell-1}, \\
    \mathbf{z}_\ell &= \text{MLP}(\text{LN}(\mathbf{z}'_\ell)) + \mathbf{z}'_\ell,
    \label{eq:msa_mlp}
\end{aligned}
\end{equation}
where LN denotes Layer Normalization and $\mathbf{z}_\ell \in \mathbb{R}^{N \times D}$ contains refined patch representations at block $\ell$. The MSA mechanism at block $\ell$ computes query $\mathbf{Q}_\ell$, key $\mathbf{K}_\ell$, and value $\mathbf{V}_\ell$ matrices from the input, then applies scaled dot-product attention across $H$ heads:
\begin{equation}
    \text{MSA}(\mathbf{z}_{\ell-1}) = \text{Concat}(\text{head}_{\ell,1}, \ldots, \text{head}_{\ell,H})\mathbf{W}^O_\ell,
\end{equation}
where each attention head $h \in \{1, \ldots, H\}$ at block $\ell$ computes: \vspace{-2mm}
\begin{equation}
    \text{head}_{\ell,h} = \text{softmax}\left(\frac{\mathbf{Q}_{\ell,h}\mathbf{K}_{\ell,h}^\top}{\sqrt{D/H}}\right)\mathbf{V}_{\ell,h},
    \label{eq:attention}
\end{equation}
with $\mathbf{Q}_{\ell,h}, \mathbf{K}_{\ell,h}, \mathbf{V}_{\ell,h} \in \mathbb{R}^{N \times (D/H)}$ and $\mathbf{W}^O_\ell \in \mathbb{R}^{D \times D}$ as the output projection matrix. The queries, keys, and values are computed as $\mathbf{Q}_{\ell,h} = \mathbf{z}_{\ell-1}\mathbf{W}^Q_{\ell,h}$, $\mathbf{K}_{\ell,h} = \mathbf{z}_{\ell-1}\mathbf{W}^K_{\ell,h}$, and $\mathbf{V}_{\ell,h} = \mathbf{z}_{\ell-1}\mathbf{W}^V_{\ell,h}$, where $\mathbf{W}^Q_{\ell,h}, \mathbf{W}^K_{\ell,h}, \mathbf{W}^V_{\ell,h} \in \mathbb{R}^{D \times (D/H)}$ are learnable projection matrices for head $h$ at block $\ell$.

\subsection{Pre-Softmax Attention for Quality Assessment}
\label{subsec:presoftmax_attention}

The attention mechanism in Equation \ref{eq:attention} computes a post-softmax attention matrix $\mathbf{A}_{\ell,h} = \text{softmax}\left(\mathbf{Q}_{\ell,h}\mathbf{K}_{\ell,h}^\top \big/ \sqrt{D/H}\right) \in \mathbb{R}^{N \times N}$, where $\mathbf{A}_{\ell,h}^{(i,j)}$ represents the normalized attention weight from patch $i$ to patch $j$ for head $h$ at block $\ell$. However, the softmax normalization constrains these values to a probability distribution, which may obscure the raw strength of relationships between patches. The pre-softmax attention scores, computed before applying the softmax function, preserve the magnitude of these relationships and may contain richer discriminative information for quality assessment. We define the pre-softmax attention matrix at block $\ell$ and head $h$ as:
\begin{equation}
    \mathbf{A}^{\text{raw}}_{\ell,h} = \frac{\mathbf{Q}_{\ell,h}\mathbf{K}_{\ell,h}^\top}{\sqrt{D/H}} \in \mathbb{R}^{N \times N},
    \label{eq:presoftmax}
\end{equation}
where $\mathbf{A}^{\text{raw}}_{\ell,h}$ contains the scaled dot-product similarities between all pairs of patches before normalization, and the scaling factor $\sqrt{D/H}$ prevents the dot products from growing too large as the embedding dimension increases \cite{DBLP:conf/nips/VaswaniSPUJGKP17}. These raw attention scores capture the unnormalized affinity between patches, where higher absolute values indicate stronger relationships (either positive or negative) between corresponding facial regions.

Our approach is grounded in the hypothesis that pre-softmax attention scores intrinsically reflect the quality of the input-model interaction, serving as a proxy for face image quality. This hypothesis is motivated by two key observations from recent ViT research: (1) attention patterns naturally encode spatial importance as ViTs inherently function as saliency learners, and (2) attention statistics capture intrinsic properties of input-model interaction, successfully serving as utility proxies for tasks like out-of-distribution detection and quality-aware recognition. Specifically, we posit that high-quality face images enable strong query-key alignments across identity-relevant facial regions, yielding attention scores with higher magnitudes and focused distributions. In contrast, quality degradations introduce patch ambiguity that weakens these alignments, producing lower-magnitude, diffuse attention patterns. Unlike post-softmax normalization, pre-softmax scores preserve these magnitude differences, making them suitable for quantifying the confidence and clarity of the model's internal facial region associations. We empirically validate this hypothesis through comprehensive evaluation on multiple benchmarks in Section \ref{sec:results}.

\begin{figure*}[t!]
    \centering
    \includegraphics[width=0.95\textwidth]{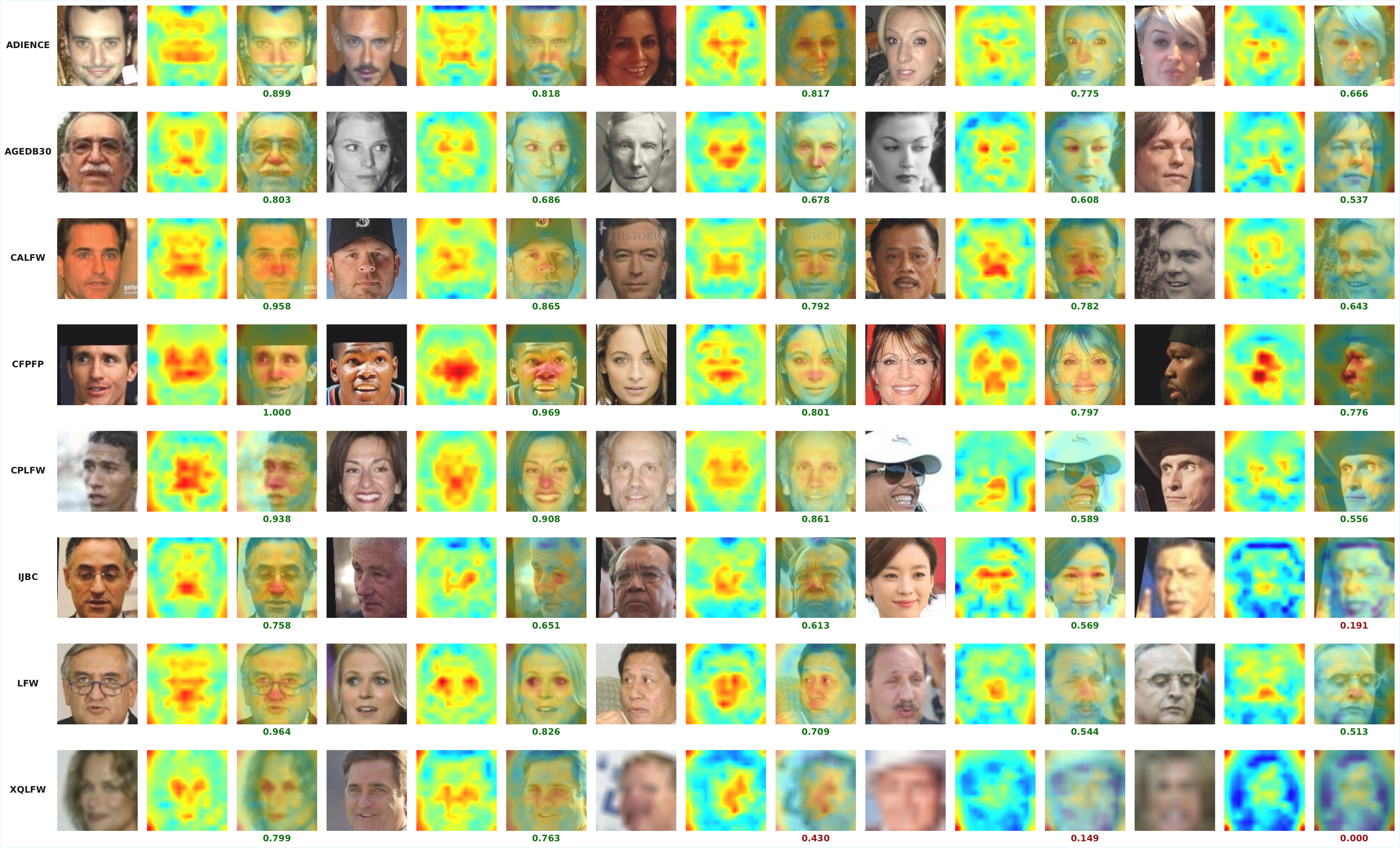}
    \caption{Multi-dataset attention analysis using ViT-S/WebFace4M/AdaFace across eight benchmark datasets (40 images total: 8 rows × 5 images per row). Each row represents one dataset, with images sorted left-to-right by quality within each row (leftmost=highest, rightmost=lowest). For each image: three columns show original, pre-softmax attention heatmap, and overlay with normalized score. All 40 images share global quality score normalization [0,1] and global attention colormap (\textcolor{myblue}{blue}=low magnitude, \textcolor{myred}{red}=high magnitude). High-quality images consistently show higher attentions, while challenging conditions (occlusions, extreme poses, low resolution) produce lower attentions regardless of the dataset source.}
    \label{fig:datasets_vits_adaface} \vspace{-6mm}
\end{figure*}

\subsection{Attention-Based Quality Score Computation}
\label{subsec:quality_computation}

To generate an image-level quality score, we propose a straightforward aggregation strategy that leverages information from all attention heads and patches in the final transformer block $\ell = L$. We focus on the final block as it contains the most refined representations after processing through all $L$ transformer layers, capturing high-level semantic relationships that are most relevant for the FR task.

For the final block $L$, we collect the pre-softmax attention matrices from all $H$ heads: $\{\mathbf{A}^{\text{raw}}_{L,1}, \mathbf{A}^{\text{raw}}_{L,2}, \ldots, \mathbf{A}^{\text{raw}}_{L,H}\}$, where each $\mathbf{A}^{\text{raw}}_{L,h} \in \mathbb{R}^{N \times N}$ captures pairwise patch relationships for head $h$. We flatten these attention matrices into a single vector that concatenates all attention scores across all heads and all patch pairs:
\begin{equation}
    \mathbf{v} = \text{Flatten}(\{\mathbf{A}^{\text{raw}}_{L,1}, \ldots, \mathbf{A}^{\text{raw}}_{L,H}\}) \in \mathbb{R}^{H \cdot N^2},
    \label{eq:flatten_attention}
\end{equation}
where $\mathbf{v}$ contains all $H \cdot N^2$ pre-softmax attention values, representing the complete attention distribution across all heads and patch interactions. This flattening operation preserves all attention information while enabling efficient aggregation.

We then compute the image-level quality score by averaging all attention values: \vspace{-2mm}
\begin{equation}
    Q = \frac{1}{H \cdot N^2} \sum_{i=1}^{H \cdot N^2} \mathbf{v}^{(i)},
    \label{eq:image_quality}
\end{equation}
where $Q \in \mathbb{R}$ is the final quality score for $\mathbf{I}$. This formulation is grounded in the observation that high-quality face images with discriminative features enable the model to establish strong query-key alignments across all attention heads, resulting in higher average attention magnitudes. Conversely, degraded images with e.g., blur or occlusions produce weaker alignments and lower average attention scores across the network. By aggregating over all attention heads and patch interactions, our approach captures the global quality signal encoded in the model's attention patterns.

\subsection{Interpretability Through Attention Visualization}
\label{subsec:interpretability}

A potential advantage of this attention-based approach is its inherent interpretability. Unlike methods that produce opaque scalar quality scores \cite{MagFace,PFE_FIQA,boutros_2023_crfiqa}, our approach provides insights into the model's attention patterns. While our quality score computation aggregates attention values globally across all heads and patches, the underlying attention matrices $\mathbf{A}^{\text{raw}}_{L,h}$ for different heads $h \in \{1, \ldots, H\}$ can be analyzed to understand which facial regions receive strong attention and how quality degradation affects these patterns. By visualizing individual attention heads or computing patch-level attention statistics from the raw attention matrices, practitioners can identify which facial regions contribute to quality assessment. For instance, occlusions may manifest as reduced attention involving the occluded regions, while blur may result in diffuse, weakened attention patterns across the entire face. This capability provides valuable diagnostic information for quality control in biometric systems, enabling targeted image acquisition improvements or informed decisions about sample retention or rejection.


\begin{figure}[t]
    \centering
    \setlength{\tabcolsep}{1pt}
    \renewcommand{\arraystretch}{1.1}
    \resizebox{0.95\linewidth}{!}{
        \includegraphics[width=0.5\textwidth]{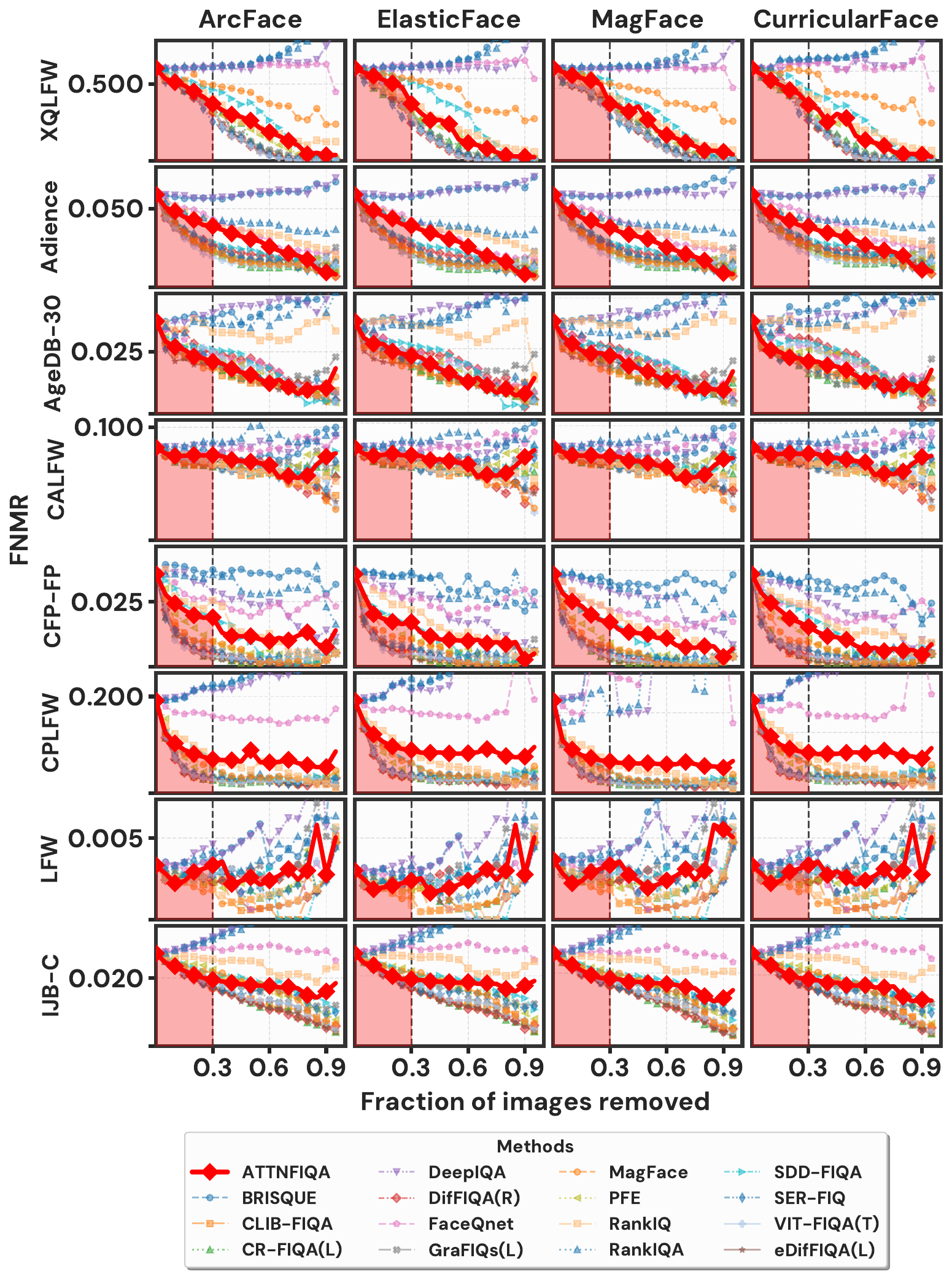}}
    \caption{EDCs for FNMR@FMR=$1e-3$ of our proposed \ourmethod in comparison to SOTA. Results shown on eight benchmark datasets: LFW , AgeDB-30, CFP-FP, CALFW, Adience, CPLFW, XQLFW, and IJB-C, using ArcFace, ElasticFace, MagFace, and CurricularFace FR models. The proposed \ourmethod is marked with {\color{red}solid red} lines.}
    \label{fig:fnmr3} \vspace{-6mm}
\end{figure}

\section{IMPLEMENTATION DETAILS}

\textbf{Pre-trained Models:} We evaluate \ourmethod using three different pre-trained ViT-based FR models with different architectures and also trained with different loss functions: ViT-B/WebFace4M/AdaFace \cite{DBLP:conf/cvpr/Kim0L22}, ViT-S/WebFace4M/AdaFace \cite{DBLP:conf/cvpr/Kim0L22} and ViT-S/WebFace4M/ArcFace \cite{DBLP:conf/cvpr/Kim0L22}. ViT-B model utilizes 24 transformer blocks, and 8 attention heads per block. Both ViT-S models utilize the architecture with 12 transformer blocks, 8 attention heads per block ($H=8$). All models have an embedding dimension of $D=512$, and were trained on the WebFace4M  \cite{DBLP:journals/pami/ZhuHDYHCZYDLZ23}. The input face images are aligned and cropped to $112 \times 112$ pixels, resulting in $N = 144$ patches with patch size $P = 8$. These publicly available models provide a strong baseline for evaluating our training-free FIQA approach across different architectures (ViT-B vs. ViT-S) and different margin-based loss functions (AdaFace vs. ArcFace) while maintaining consistent training data. For each input face image, we perform a single forward pass through the pre-trained ViT and extract the pre-softmax attention matrices $\{\mathbf{A}^{\text{raw}}_{L,h}\}_{h=1}^{H}$ from the final transformer block ($L=12$ for ViT-S). Following Equation \ref{eq:flatten_attention}, we flatten these matrices into a single vector containing all $H \cdot N^2$ attention values, then compute the image-level quality score via Equation \ref{eq:image_quality} by averaging across all values. No additional training, fine-tuning, or architectural modifications are required, enabling immediate deployment on any pre-trained ViT-based FR model with minimal computational overhead.

\textbf{Evaluation Benchmarks and Metrics:} We assess the effectiveness of \ourmethod on eight benchmark datasets: Labeled Faces in the Wild (LFW) \cite{LFWTech}, AgeDB-30 \cite{agedb}, Celebrities in Frontal-Profile in the Wild (CFP-FP) \cite{cfp-fp}, Cross-Age LFW (CALFW) \cite{CALFW}, Adience \cite{Adience}, Cross-Pose LFW (CPLFW) \cite{CPLFWTech}, Cross-Quality LFW (XQLFW) \cite{XQLFW}, and IJB-C \cite{ijbc}. 
Performance is measured using Error-versus-Discard Characteristic (EDC) curves \cite{GT07}, which assess the impact of discarding low-quality face images on face verification performance and quantify how verification errors decrease as low-quality samples are progressively removed. The False Non-Match Rate (FNMR) is evaluated at fixed False Match Rate (FMR) thresholds \cite{iso_metric}, specifically at $1e-3$ (recommended for border control by Frontex \cite{frontex2015best}) and $1e-4$ (for higher security applications). Additionally, we report the Area Under the Curve (AUC) and partial AUC (pAUC) of the EDC curves to quantify verification performance across rejection rates. The pAUC measures performance up to a 30\% rejection rate following established protocols \cite{10449044, babnikTBIOM2024, DBLP:journals/tbbis/SchlettRTB24}. To thoroughly examine the impact of our FIQA approach across different FR architectures, we evaluate performance using four FR models (ArcFace \cite{deng2019arcface}, ElasticFace (ElasticFace-Arc) \cite{elasticface}, MagFace \cite{MagFace}, CurricularFace \cite{curricularFace}). Each model processes $112 \times 112$ aligned images to generate 512-dimensional feature embeddings. We use officially released models by each FR model. All evaluations are conducted under cross-model settings, where the models used to evaluate FIQA (ViT-based models) are different from those used to extract face feature representations (CNN-based models), demonstrating the generalizability of our approach across different architectures.

\textbf{Empirical Validation on SynFIQA:} To empirically validate our hypothesis that pre-softmax attention scores correlate with face image quality, we analyze the SynFIQA dataset \cite{mrfiqa}, a quality-labeled synthetic dataset containing approximately 550,000 images generated through a two-stage pipeline based on stable diffusion with controllable 3D facial parameters, dual text prompts for occlusions, and post-processing for blur and downsampling. The dataset comprises 5,000 identities, each with 10 reference images and 100 degraded variants (10 per reference), organized into 11 quality groups: Q0-Q9 represent degraded images progressing from lowest to highest quality, while the Ref group contains the original high-quality reference images. For each image, we compute \ourmethod scores using our proposed method and analyze their distribution across quality groups. As shown in Figure \ref{fig:synfiqa_validation}, the boxplot visualization reveals a clear quality gradient: Q0 (lowest quality) exhibits the lowest mean \ourmethod score, which progressively increases through Q1-Q9, with Ref (reference images) achieving the highest scores. This monotonic trend demonstrates that our attention-based quality scores effectively distinguish between quality groups. The statistical analysis shows increasing mean scores across groups.


\begin{table*}[!t]
\centering
\setlength{\tabcolsep}{3pt}
\vspace{1mm}
\caption{Ablation studies, metric being pAUC-EDC (the lower, the better), for \ourmethod investigating different configurations: architecture depth (ViT-S vs ViT-B), training loss (AdaFace vs ArcFace), head aggregation strategies (concat all vs individual heads), and aggregation metrics (mean, max, median, 1/std). Mean pAUC-EDC computed across seven benchmarks at FMR=$1e-3$ and $1e-4$. Best per study in \textbf{bold}. All values are scaled by a $10^3$ factor for better readability.
\vspace{-1mm}}
\label{tab:ablation_pauc_attnfiqa}
\resizebox{\textwidth}{!}{
\begin{tabular}{|c|c|
cc|cc|cc|cc|cc|cc|cc|cc|}
\hline
\multirow{2}{*}{\textbf{Study}} & \multirow{2}{*}{\textbf{Method}}
& \multicolumn{2}{c|}{Adience \cite{Adience}} 
& \multicolumn{2}{c|}{AgeDB-30 \cite{agedb}} 
& \multicolumn{2}{c|}{CFP-FP \cite{cfp-fp}} 
& \multicolumn{2}{c|}{LFW \cite{LFWTech}} 
& \multicolumn{2}{c|}{CALFW \cite{CALFW}} 
& \multicolumn{2}{c|}{CPLFW \cite{CPLFWTech}} 
& \multicolumn{2}{c|}{XQLFW \cite{XQLFW}} 
& \multicolumn{2}{c|}{Mean pAUC-EDC} \\

\cline{3-18}
 & 
 & $1e{-3}$ & $1e{-4}$ 
 & $1e{-3}$ & $1e{-4}$ 
 & $1e{-3}$ & $1e{-4}$ 
 & $1e{-3}$ & $1e{-4}$ 
 & $1e{-3}$ & $1e{-4}$ 
 & $1e{-3}$ & $1e{-4}$ 
 & $1e{-3}$ & $1e{-4}$ 
 & $1e{-3}$ & $1e{-4}$ \\

\hline
\multirow{2}{*}{Architecture}
 & ViT-S/WebFace4M/AdaFace
 & 13.7643 &	34.0544	&\textbf{7.8214}	&\textbf{11.1299}&	\textbf{7.0003}	&\textbf{8.9304}&	\textbf{0.8396}	&\textbf{1.0232}	&\textbf{22.7711}	&\textbf{24.7382}	&\textbf{30.1673}&	\textbf{44.5971}	&\textbf{144.4323}&	\textbf{178.3813}& \textbf{32.3995}	&\textbf{43.2649} \\
 & ViT-B/WebFace4M/AdaFace
 &\textbf{12.8261}&	\textbf{30.5972} &9.9527	&15.2625	&8.8310&	13.1429&	1.0190&	1.1987&	23.9418&	25.7059&	50.0956&	68.2659&	176.2205	&196.2156& 40.4124 &	50.0555 \\
\hline

\multirow{2}{*}{Loss Function}
 & ViT-S/WebFace4M/AdaFace
 & \textbf{13.7643} &	\textbf{34.0544	}&\textbf{7.8214}	&\textbf{11.1299}&	7.0003	&\textbf{8.9304}&	0.8396	&\textbf{1.0232}	&22.7711	&\textbf{24.7382}	&\textbf{30.1673}&	\textbf{44.5971}	&\textbf{144.4323}&	\textbf{178.3813}& \textbf{32.3995}	&\textbf{43.2649 }\\
\hline
\end{tabular}}\vspace{-3mm}
\end{table*}

\begin{table*}[ht!]
    \begin{center}
    \vspace{1mm}
    \caption{The pAUC-EDC (the lower, the better) achieved by \ourmethod and the SOTA methods under different experimental settings. The notions of $1e-3$ and $1e-4$ indicate the value of the fixed FMR at which the EDC curves (FNMR vs. reject) were calculated. All values are scaled by a $10^3$ factor for better readability.  
    \vspace{-1mm}
    }
    \label{tab:sota_pauc}
    \resizebox{\textwidth}{!}{
\begin{tabular}{|c|cl|cc|cc|cc|cc|cc|cc|cc|cc|}
    \hline
\multirow{3}{*}{FR} & \multicolumn{2}{c|}{\multirow{3}{*}{Method}}
& \multicolumn{2}{c|}{Adience} & \multicolumn{2}{c|}{AgeDB-30} & \multicolumn{2}{c|}{CFP-FP}
& \multicolumn{2}{c|}{LFW} & \multicolumn{2}{c|}{CALFW} & \multicolumn{2}{c|}{CPLFW}
& \multicolumn{2}{c|}{XQLFW} & \multicolumn{2}{c|}{IJB-C} \\
 & &
 & \multicolumn{2}{c|}{\cite{Adience}} & \multicolumn{2}{c|}{\cite{agedb}} & \multicolumn{2}{c|}{\cite{cfp-fp}}
 & \multicolumn{2}{c|}{\cite{LFWTech}} & \multicolumn{2}{c|}{\cite{CALFW}} & \multicolumn{2}{c|}{\cite{CPLFWTech}}
 & \multicolumn{2}{c|}{\cite{XQLFW}}  & \multicolumn{2}{c|}{\cite{ijbc}} \\ \cline{4-19}
                                                                                    &                                                   &  & $1e{-3}$ & $1e{-4}$ & $1e{-3}$ & $1e{-4}$  & $1e{-3}$ & $1e{-4}$  & $1e{-3}$ & $1e{-4}$  & $1e{-3}$ & $1e{-4}$ & $1e{-3}$ & $1e{-4}$ & $1e{-3}$ & $1e{-4}$ & $1e{-3}$ & $1e{-4}$\\ \hline
     \hline
     \multirow{17}{*}{\rotatebox[origin=c]{90}{ArcFace\cite{deng2019arcface}}}      & \multirow{3}{*}{\rotatebox[origin=c]{90}{IQA}}    & \multicolumn{1}{|l|}{BRISQUE\cite{BRISQE_IQA}} &17.1746 &39.8035 &11.5267 &17.4919 &11.3476 &16.2225 &1.0438 &1.2273 &23.9653 &26.8772 &61.1376 &77.2797 &181.8562 &203.0627 &8.7528 &13.7855                                                \\
                                                                                    &                                                   & \multicolumn{1}{|l|}{RankIQA\cite{liu2017rankiqa}} &14.6239 &35.4490 &10.5402 &17.0561 &10.5550 &15.9389 &1.0739 &1.2560 &25.2499 &28.2475 &61.6504 &80.0431 &184.0819 &205.2308 &8.6993 &13.6493                                                \\
\hline
    \end{tabular}}
    \end{center}
    \end{table*}

\section{RESULTS}
\label{sec:results}


\subsection{Qualitative Analysis: Attention Visualization}
\label{subsec:qualitative}

To provide interpretability and validate that attention patterns correlate with image quality factors, we visualize pre-softmax attention maps from pre-trained ViT models across different scenarios. We examine both controlled degradation conditions (single subject with various quality factors) and cross-dataset comparisons (diverse subjects and capture conditions). All visualizations use a global colormap applied across all images in each figure, where \textcolor{myblue}{blue} indicates lower pre-softmax attention magnitudes (correlating with lower quality) and \textcolor{myred}{red} indicates higher magnitudes (correlating with higher quality). Quality scores are normalized to [0,1] range within each figure for relative comparison. 

Figure \ref{fig:controlled_vits_adaface} shows attention maps for controlled conditions where a single face is subjected to various degradations. Images are sorted by predicted quality score from top-left (highest) to bottom-right (lowest). For each image, we display three columns: (1) original image with condition label, (2) attention heatmap showing spatial importance with global colormap normalization, and (3) overlay visualization with normalized quality score. The visualizations reveal that high-quality frontal poses exhibit focused, high-magnitude attention (\textcolor{myred}{red} heatmaps) on discriminative facial regions, resulting in high quality scores. In contrast, degraded images with occlusions, extreme poses, or face covers show diffuse, low-magnitude attention patterns (\textcolor{myblue}{blue} heatmaps), reflecting the model's uncertainty and resulting in low quality scores. 

Figure \ref{fig:datasets_vits_adaface} presents cross-dataset attention analysis across eight benchmark datasets (40 images total: 8 datasets × 5 images per dataset). Each row represents a different dataset, with images within each row sorted by quality from left (highest) to right (lowest). All 40 images share a global colormap normalization and quality score normalization [0,1]. This visualization demonstrates how attention patterns generalize across diverse data distributions, consistently showing high-magnitude attention (\textcolor{myred}{red}) on identity-relevant facial features in high-quality samples and low-magnitude attention (\textcolor{myblue}{blue}) in challenging scenarios. 

These visualizations demonstrate three key findings: (1) pre-softmax attention magnitudes naturally encode quality-relevant information, (2) spatial attention patterns provide interpretable explanations for quality predictions, revealing which facial regions contribute to overall quality. 

\subsection{Ablation and Design Choice Analysis}

\label{subsec:ablation}

To thoroughly investigate the design choices of \ourmethod and prove the generalization over different components, we conduct comprehensive ablation studies examining four key aspects: (1) architecture depth (ViT-S vs ViT-B), (2) training loss functions (AdaFace vs ArcFace), (3) aggregation strategies across attention heads, and (4) aggregation metrics for computing quality scores. All ablation experiments are evaluated across seven benchmark datasets (Adience, AgeDB-30, CFP-FP, LFW, CALFW, CPLFW, XQLFW), excluding IJB-C, using both pAUC-EDC and AUC-EDC metrics at FMR thresholds of $1e-3$ and $1e-4$. pAUC-EDC results are presented in Tables \ref{tab:ablation_pauc_attnfiqa}. \textbf{Architecture Depth:} Comparing ViT-S (12 blocks) and ViT-B (24 blocks) architectures, both trained on WebFace4M with AdaFace loss. ViT-S achieves lower (better) error rates on most individual benchmarks (e.g., AgeDB-30: 7.82/11.13 vs 9.95/15.26, CPLFW: 30.17/44.60 vs 50.10/68.27). \textbf{Loss Function:} The comparison between AdaFace and ArcFace loss functions using identical ViT-S/WebFace4M architecture shows that AdaFace achieves lower (better) mean pAUC-EDC scores (32.40/43.26 vs 35.45/48.11). This suggests that the adaptive margin strategy in AdaFace produces attention patterns that encode quality information more effectively than ArcFace's fixed margin approach. However, the performance gap is relatively modest across most benchmarks, indicating that pre-softmax attention magnitudes capture quality-relevant information regardless of the specific margin-based loss function used during training. \textbf{Head Aggregation Strategy:} The head aggregation study reveals that concatenating all attention heads and computing the mean across all values achieves the best performance with the lowest error rates (32.40/43.26), though, averaging the individual head outputs produces similar results. Individual attention heads show varying discriminative power: Head 5 and 7 performs better than the others, while Head 2 performs the worst. \textbf{Aggregation Metric:} Comparing different aggregation metrics for computing quality scores from flattened attention values shows that mean aggregation achieves the lowest (best) pAUC-EDC scores (32.40/43.26), outperforming maximum (40.35/51.60), median (33.61/43.95), and inverse standard deviation (34.22/45.24).

\subsection{Comparison to State-of-the-Art}
\label{subsec:sota}

To investigate the degree to which pre-softmax attention scores can serve as face image quality proxies, we evaluate \ourmethod against 15 IQA and FIQA methods across four CNN-based FR models (ArcFace, ElasticFace, MagFace, CurricularFace) and eight benchmark datasets. Results are presented in Table \ref{tab:sota_pauc} using pAUC-EDC metrics (lower is better). Our analysis reveals that while \ourmethod achieves competitive performance on constrained benchmarks with specific quality degradation patterns (age variations in AgeDB-30/CALFW, pose variations in CFP-FP/CPLFW, quality variations in XQLFW), the method demonstrates particularly strong effectiveness on unconstrained scenarios. On IJB-C, a large-scale benchmark containing diverse real-world quality variations without specific degradation patterns, \ourmethod achieves 6.74/10.28 pAUC-EDC at FMR $1e-3$/$1e-4$ for ArcFace, 6.49/10.00 for ElasticFace, and 6.46/9.44 for CurricularFace, performing comparably to specialized FIQA methods like ViT-FIQA (6.56/10.12, 6.33/9.76, 6.34/9.32) and outperforming several established approaches. This suggests that when image sets contain specific property variations (uniform age ranges, controlled pose angles), attention magnitudes may struggle to differentiate fine-grained quality differences within those constrained distributions. However, in unconstrained settings with diverse quality factors, the attention-based approach effectively captures the broader quality spectrum, confirming that pre-softmax attention scores serve as meaningful quality proxies particularly for real-world deployment scenarios where quality degradations are heterogeneous rather than systematically controlled.


\section{CONCLUSION}

We demonstrated that pre-softmax attention scores from pre-trained ViT-based FR models effectively serve as face image quality proxies without additional training or architectural modifications. Evaluation across eight benchmarks confirms that attention magnitudes intrinsically encode quality: high-quality images produce focused, high-magnitude patterns while degraded images generate diffuse, low-magnitude patterns. The approach provides spatial interpretability, revealing which facial regions contribute to quality determination, a critical advantage over existing methods producing opaque scalar scores. Empirical studies show ViT-S outperforms ViT-B, AdaFace slightly better than ArcFace, and mean aggregation optimal for quality estimation. The method achieves competitive performance on constrained benchmarks and particularly strong results on unconstrained IJB-C. Key advantage is immediate deployability to any ViT-based FR system can extract quality scores during standard forward passes without separate FIQA modules.


\section*{ETHICAL IMPACT STATEMENT}
Our FIQA research improves biometric system reliability but carries ethical risks: biased quality assessments may disadvantage certain demographic groups, enabling discriminatory service access or more effective mass surveillance. Our reliance on pre-trained ViT attention mechanisms introduces additional concern, as poorly understood biases in these mechanisms could propagate to quality predictions and affect demographic groups unevenly. To mitigate this, we advocate evaluating FR models on diverse demographic datasets, conducting regular bias audits across recognition and quality predictions, ensuring deployment within legal frameworks (e.g., GDPR, BIPA) with proper oversight and user consent, and including human review in high-stakes automated decisions. These extend to the data level as well: recognizing that authentic face data is inherently sensitive and its collection or misuse can itself constitute a privacy harm, we support and employ privacy-preserving alternatives such as synthetic data, as reflected in our empirical evaluation. We encourage equitable, transparent FIQA development and reject all malicious or illegal applications of this work.

\clearpage
{\small
\bibliographystyle{ieee}
\bibliography{egbib}
}

\clearpage
\appendix

\begin{figure*}[p]
    \centering
    \includegraphics[width=0.95\textwidth]{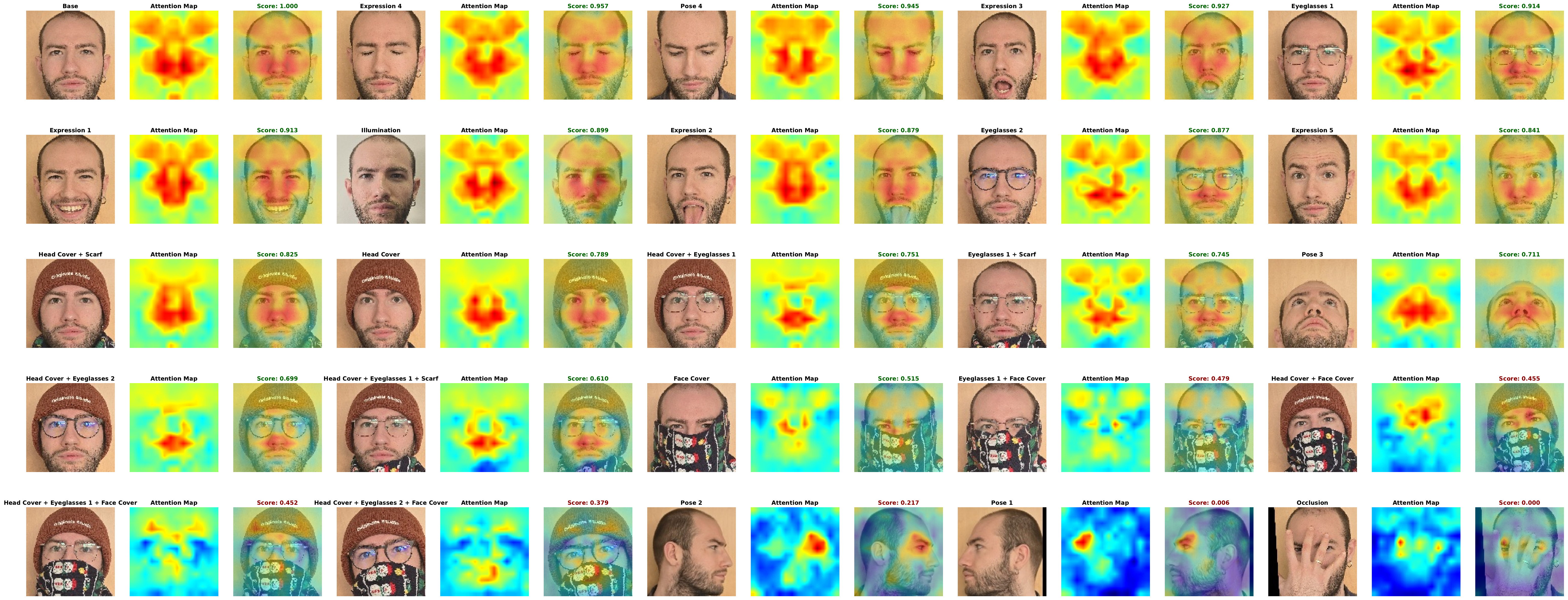}
    \caption{Attention visualization for controlled quality conditions using ViT-S/WebFace4M/ArcFace model. Despite different training objective (ArcFace vs. AdaFace), attention patterns similarly encode quality information.}
    \label{fig:controlled_vits_arcface}
\end{figure*}

\begin{figure*}[p]
    \centering
    \includegraphics[width=0.95\textwidth]{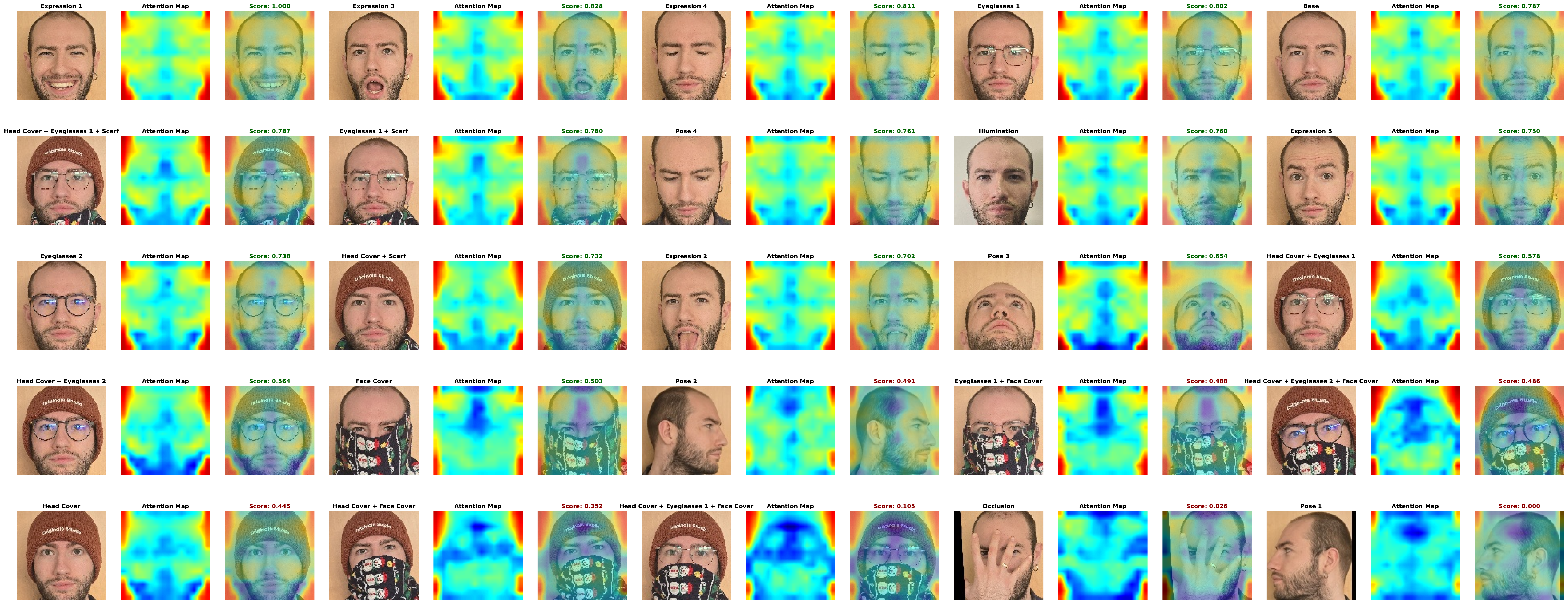}
    \caption{Attention visualization for controlled quality conditions using ViT-B/WebFace4M/AdaFace model (deeper 24-block architecture). The deeper ViT-B architecture shows similar attention-quality correlation as ViT-S models: high-magnitude red attention on facial regions for high-quality frontal poses, low-magnitude blue attention for degraded conditions. This demonstrates architecture-agnostic behavior of attention-based quality assessment.}
    \label{fig:controlled_vitb_adaface}
\end{figure*}

\begin{figure*}[p]
    \centering
    \includegraphics[width=0.95\textwidth]{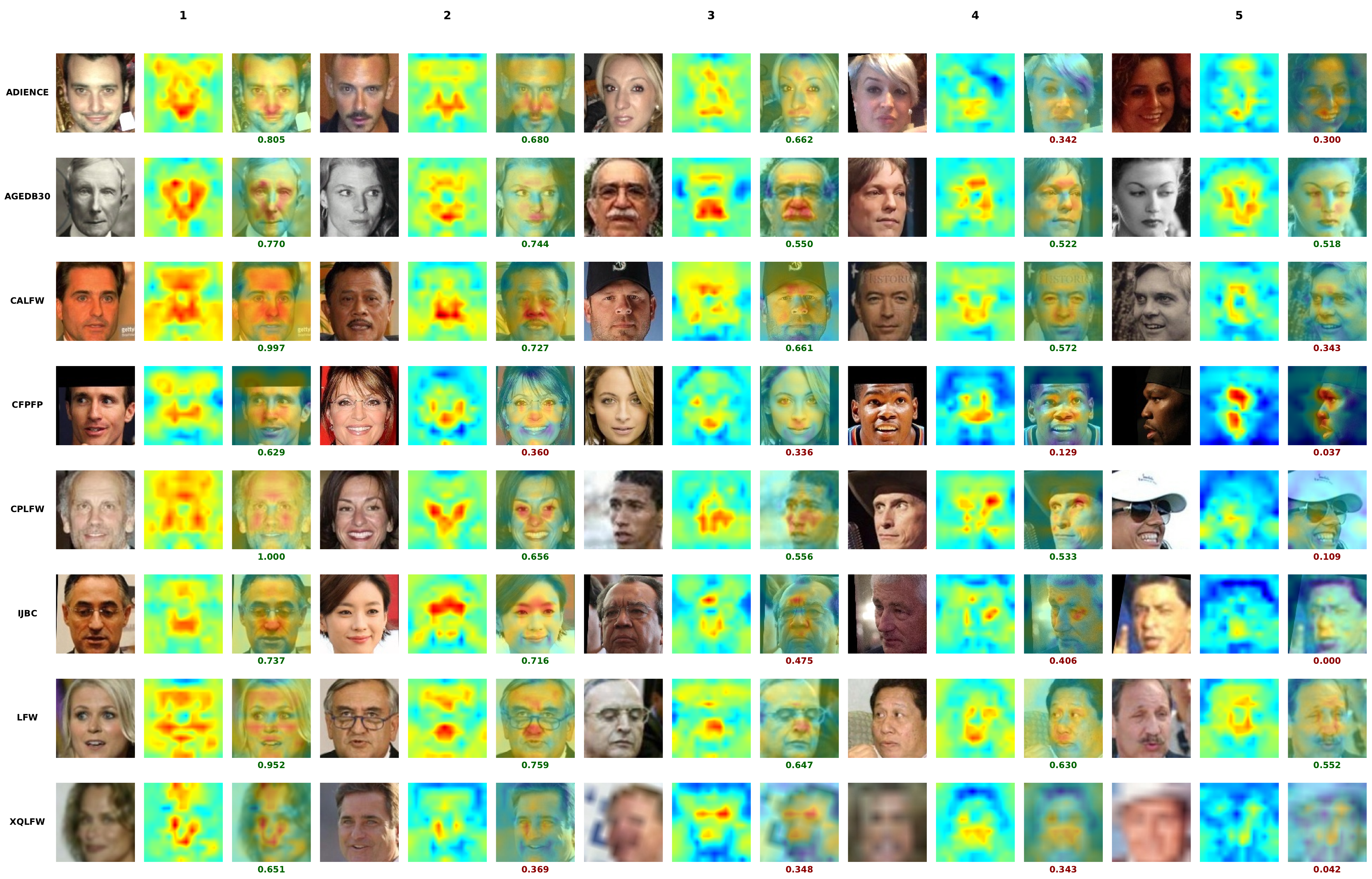}
    \caption{Multi-dataset attention analysis using ViT-S/WebFace4M/ArcFace model. Despite different training objective (ArcFace vs. AdaFace), attention patterns show similar behavior in the main paper. This demonstrates training-objective-agnostic generalization.}
    \label{fig:datasets_vits_arcface}
\end{figure*}

\begin{figure*}[p]
    \centering
    \includegraphics[width=0.95\textwidth]{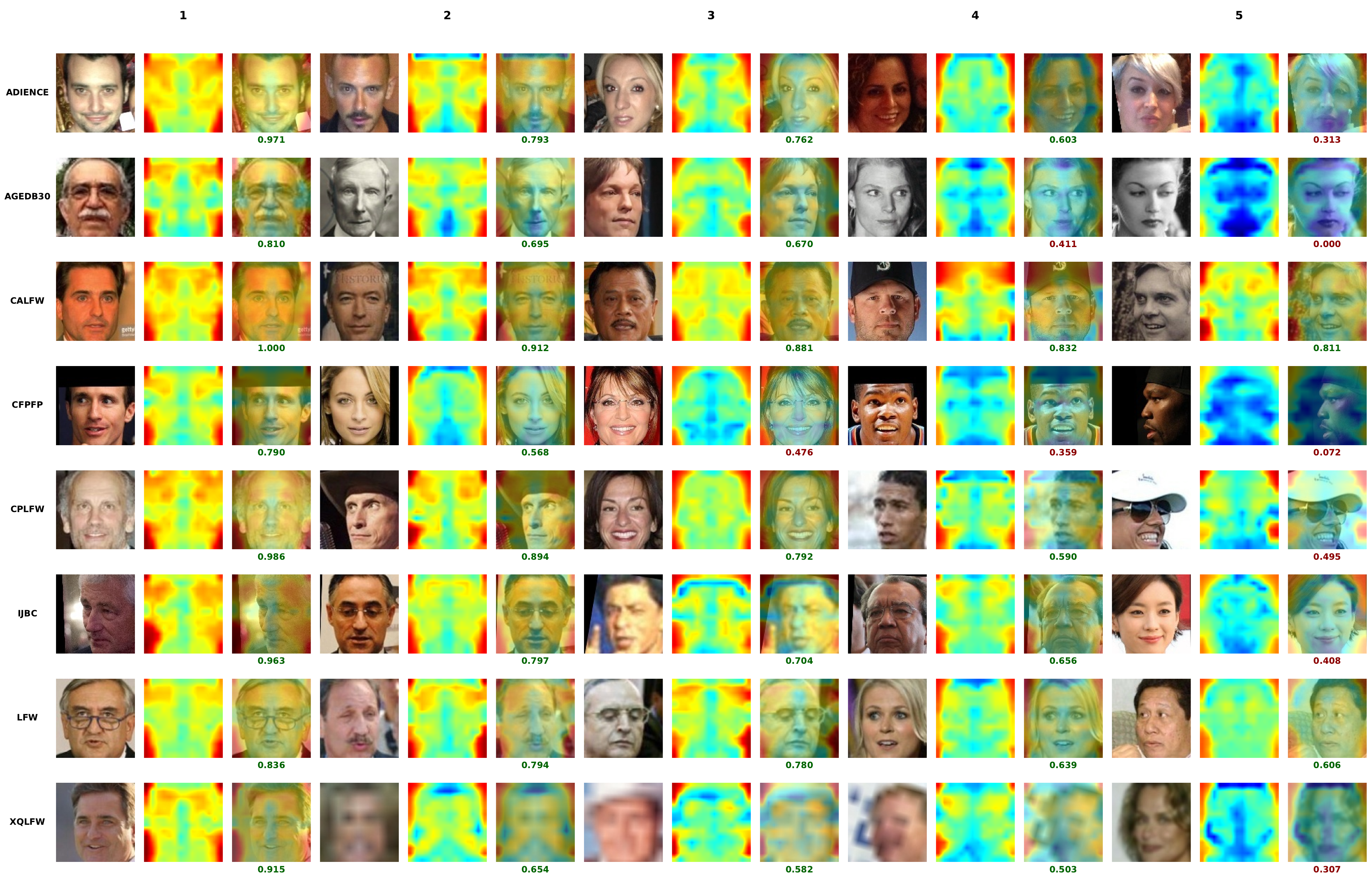}
    \caption{Multi-dataset attention analysis using ViT-B/WebFace4M/AdaFace (deeper 24-layer architecture). The deeper ViT-B model exhibits similar attention patterns as ViT-S variants: architecture depth does not fundamentally change the attention-quality relationship, confirming \ourmethod can be utilized to any ViT-based face recognition model.}
    \label{fig:datasets_vitb_adaface}
\end{figure*}

\begin{table*}[!ht]
\centering
\setlength{\tabcolsep}{3pt}
\caption{Ablation studies, metric being AUC-EDC (the lower, the better), for \ourmethod investigating different configurations: architecture depth (ViT-S vs ViT-B), training loss (AdaFace vs ArcFace), head aggregation strategies (concat all vs individual heads), and aggregation metrics (mean, max, median, 1/std). Mean AUC-EDC computed across seven benchmarks (excluding IJB-C) at FMR=$1e-3$ and $1e-4$. All values are scaled by a $10^3$ factor of for better readability.}
\label{tab:ablation_auc_attnfiqa}
\resizebox{\textwidth}{!}{
\begin{tabular}{|c|c|
cc|cc|cc|cc|cc|cc|cc|cc|}
\hline
\multirow{2}{*}{\textbf{Study}} & \multirow{2}{*}{\textbf{Method}}
& \multicolumn{2}{c|}{Adience \cite{Adience}} 
& \multicolumn{2}{c|}{AgeDB-30 \cite{agedb}} 
& \multicolumn{2}{c|}{CFP-FP \cite{cfp-fp}} 
& \multicolumn{2}{c|}{LFW \cite{LFWTech}} 
& \multicolumn{2}{c|}{CALFW \cite{CALFW}} 
& \multicolumn{2}{c|}{CPLFW \cite{CPLFWTech}} 
& \multicolumn{2}{c|}{XQLFW \cite{XQLFW}} 
& \multicolumn{2}{c|}{Mean AUC-EDC} \\

\cline{3-18}
 & 
 & $1e{-3}$ & $1e{-4}$ 
 & $1e{-3}$ & $1e{-4}$ 
 & $1e{-3}$ & $1e{-4}$ 
 & $1e{-3}$ & $1e{-4}$ 
 & $1e{-3}$ & $1e{-4}$ 
 & $1e{-3}$ & $1e{-4}$ 
 & $1e{-3}$ & $1e{-4}$ 
 & $1e{-3}$ & $1e{-4}$ \\

\hline
\multirow{2}{*}{Architecture}
 & ViT-S/WebFace4M/AdaFace
 & 29.5989	&	69.4211	&	16.804	&	20.9394	&	14.4068	&	17.9232	&	2.8015	&	3.4766	&	65.873	&	70.0654	&	73.0611	&	97.0756	&	259.914	&	314.2399			&	66.0656	&	84.7345	\\
 & ViT-B/WebFace4M/AdaFace
 & 	24.4922	 &48.2965		&	20.3944	&	28.0008	&	20.0628	&	28.5818	&	2.9957	&	3.4543	&	65.3171	&	67.6438	&	99.6102	&	136.4547	&	461.636	&	502.1325			&		99.2155	& 116.3663	\\
\hline

\multirow{2}{*}{Loss Function}
 & ViT-S/WebFace4M/AdaFace
 & 29.5989	&	69.4211	&	16.804	&	20.9394	&	14.4068	&	17.9232	&	2.8015	&	3.4766	&	65.873	&	70.0654	&	73.0611	&	97.0756	&	259.914	&	314.2399			&	66.0656	&	84.7345	\\
 & ViT-S/WebFace4M/ArcFace
 & 36.5955	&	84.3606	&	23.7328	&	32.1956	&	17.1291	&	23.6591	&	1.8267	&	2.3621	&	62.3303	&	66.7056	&	121.7123	&	147.6327	&	372.4279	&	502.2187			&	90.8221	&	122.7335	\\
\hline
\end{tabular}}
\end{table*}

\begin{table*}[ht!]
    \begin{center}
    \caption{The AUCs of EDC (the lower, the better) achieved by our \ourmethod and the SOTA methods under different experimental settings. The notions of $1e-3$ and $1e-4$ indicate the value of the fixed FMR at which the EDC curves(FNMR vs. reject) were calculated. All values are scaled by a $10^3$ factor of for better readability.}
    \label{tab:sota_auc}
    \resizebox{0.99\textwidth}{!}{
\begin{tabular}{|c|cl|cc|cc|cc|cc|cc|cc|cc|cc|}
    \hline
\multirow{3}{*}{FR} & \multicolumn{2}{c|}{\multirow{3}{*}{Method}}
& \multicolumn{2}{c|}{Adience} & \multicolumn{2}{c|}{AgeDB-30} & \multicolumn{2}{c|}{CFP-FP}
& \multicolumn{2}{c|}{LFW} & \multicolumn{2}{c|}{CALFW} & \multicolumn{2}{c|}{CPLFW}
& \multicolumn{2}{c|}{XQLFW} & \multicolumn{2}{c|}{IJB-C}\\
 & &
 & \multicolumn{2}{c|}{\cite{Adience}} & \multicolumn{2}{c|}{\cite{agedb}} & \multicolumn{2}{c|}{\cite{cfp-fp}}
 & \multicolumn{2}{c|}{\cite{LFWTech}} & \multicolumn{2}{c|}{\cite{CALFW}} & \multicolumn{2}{c|}{\cite{CPLFWTech}}
 & \multicolumn{2}{c|}{\cite{XQLFW}} & \multicolumn{2}{c|}{\cite{ijbc}} \\ \cline{4-19}
                                                                                    &                                                   &  & $1e{-3}$ & $1e{-4}$ & $1e{-3}$ & $1e{-4}$  & $1e{-3}$ & $1e{-4}$  & $1e{-3}$ & $1e{-4}$  & $1e{-3}$ & $1e{-4}$ & $1e^{-3}$ & $1e{-4}$ & $1e{-3}$ & $1e{-4}$ & $1e{-3}$ & $1e{-4}$\\ \hline
     \hline
     \multirow{17}{*}{\rotatebox[origin=c]{90}{ArcFace\cite{deng2019arcface}}}      & \multirow{3}{*}{\rotatebox[origin=c]{90}{IQA}}    & \multicolumn{1}{|l|}{BRISQUE\cite{BRISQE_IQA}} &56.5241 &128.4787 &39.9633 &58.5217 &34.2725 &43.2831 &4.2931 &4.9286 &75.4879 &81.2974 &255.8091 &303.7303 &667.9559 &712.1749 &38.1258 &65.5540                                                \\
                                                                                    &                                                   & \multicolumn{1}{|l|}{RankIQA\cite{liu2017rankiqa}} &40.0200 &93.3170 &37.1815 &52.3012 &30.1087 &38.4222 &3.9168 &4.5469 &84.6434 &91.5241 &243.6630 &296.8909 &658.3605 &703.8726 &38.5116 &63.9573                                                \\
\hline
    \end{tabular}}
    \end{center}
    \end{table*}

\begin{figure*}[t]
    \centering
    \setlength{\tabcolsep}{1pt}
    \renewcommand{\arraystretch}{1.1}
    \resizebox{0.85\textwidth}{!}{
        \includegraphics[width=0.44\textwidth]{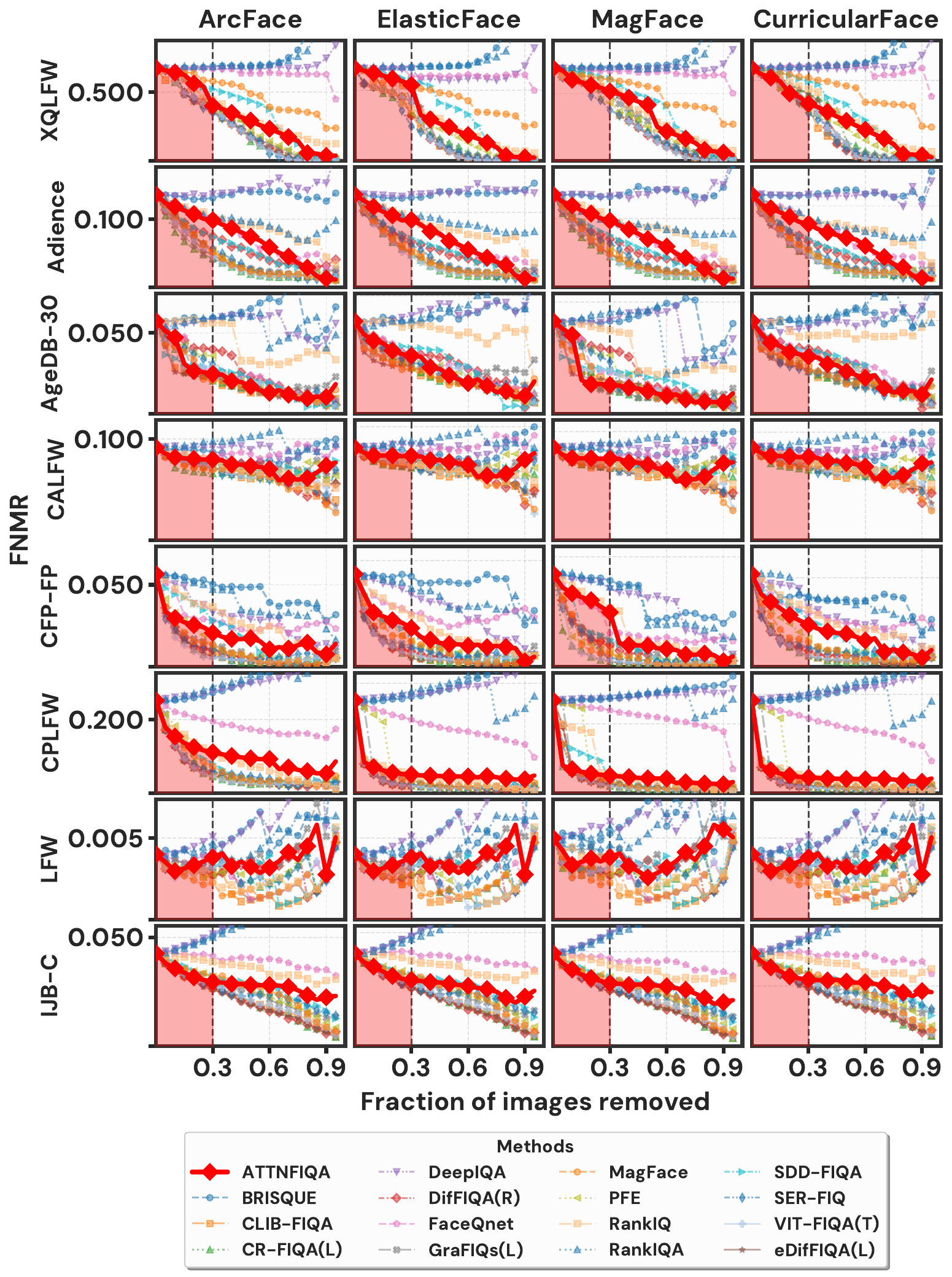}}
    \caption{Error-versus-Discard Characteristic (EDC) curves for FNMR@FMR=$1e-4$ of our proposed method \ourmethod in comparison to SOTA. Results shown on eight benchmark datasets: LFW \cite{LFWTech}, AgeDB-30 \cite{agedb}, CFP-FP \cite{cfp-fp}, CALFW \cite{CALFW}, Adience \cite{Adience}, CPLFW \cite{CPLFWTech}, XQLFW \cite{XQLFW}, and IJB-C \cite{ijbc}, using ArcFace \cite{deng2019arcface}, ElasticFace \cite{elasticface}, MagFace \cite{MagFace}, and CurricularFace \cite{curricularFace} FR models. The proposed \ourmethod are marked with {\color{red}solid red} lines, respectively.}
    \label{fig:fnmr4}
\end{figure*}

\begin{figure*}[t]
    \centering
    \setlength{\tabcolsep}{1pt}
    \renewcommand{\arraystretch}{1.1}
    \resizebox{\textwidth}{!}{
        \includegraphics[width=\textwidth]{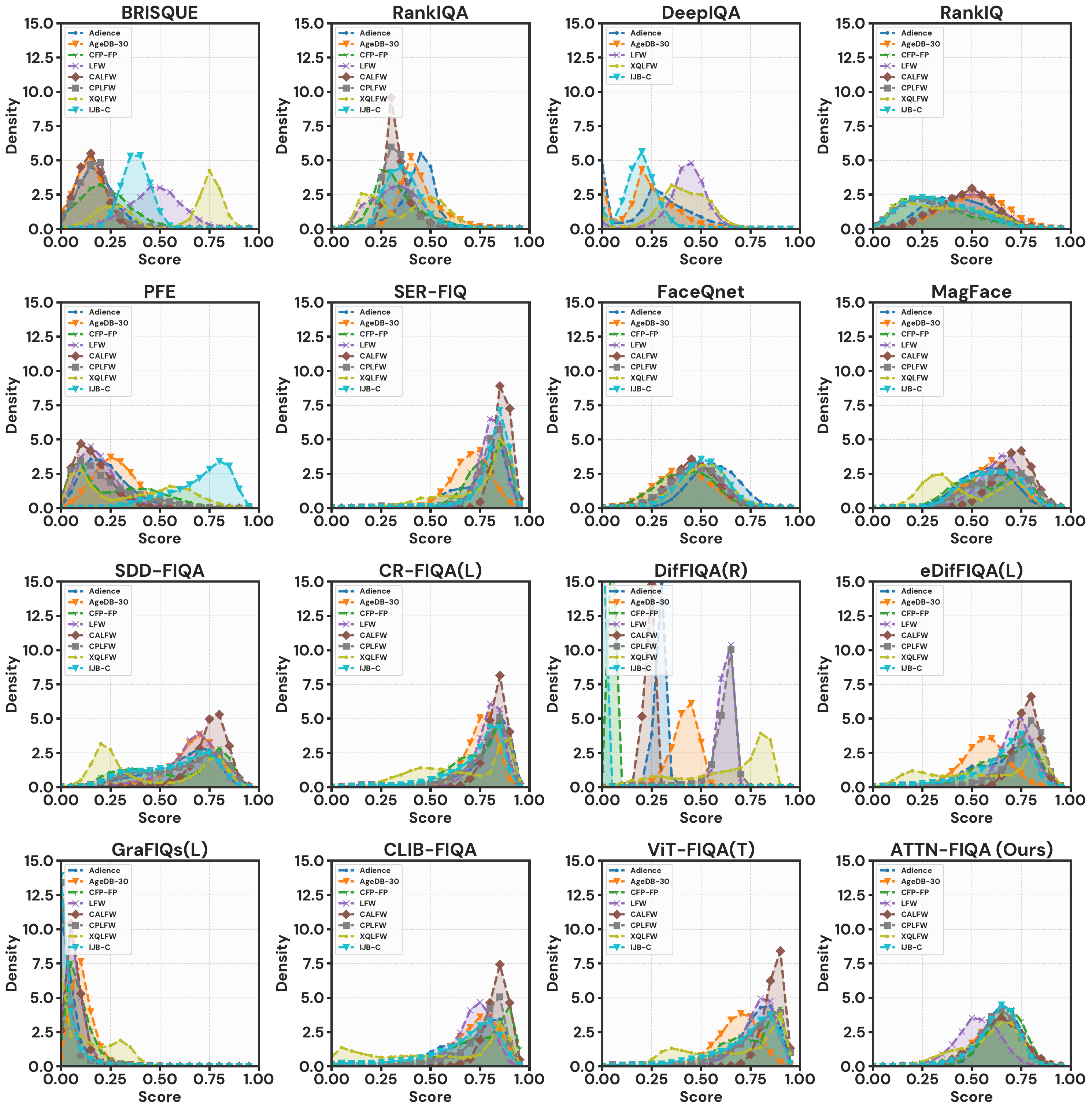}}
    \caption{Distribution of quality scores across the evaluation benchmarks, comparing our proposed method (\ourmethod) with SOTA methods. All scores are normalized to the range [0, 1].}
    \label{fig:quality_dist}
\end{figure*}

This supplementary material provides additional quantitative and visual evidence supporting the paper titled \textbf{ATTN-FIQA: Interpretable Attention-based Face Image Quality Assessment with Vision Transformers}.

\section*{Tables}

Table \ref{tab:ablation_auc_attnfiqa} presents full AUC metrics for ablation studies, evaluating: (1) architecture depth (ViT-S vs ViT-B), (2) loss function (AdaFace vs ArcFace), (3) head aggregation (individual heads vs concat vs mean), and (4) aggregation metrics (mean, max, median, 1/std). Results show concatenating all heads with mean aggregation achieves optimal performance. Table \ref{tab:sota_auc} provides complete SOTA comparison with full AUC-EDC metrics across all benchmarks and FR models.

\section*{Figures}

\subsection*{Controlled Degradation Analysis}

Figures \ref{fig:controlled_vits_arcface} and \ref{fig:controlled_vitb_adaface} show attention visualizations for controlled quality conditions. Figure \ref{fig:controlled_vits_arcface} uses ViT-S/ArcFace to assess loss function impact. Figure \ref{fig:controlled_vitb_adaface} uses ViT-B/AdaFace (24 blocks) to examine architectural depth effects. Both use three-column format: original image, attention heatmap (blue=low magnitude, red=high magnitude), and overlay with quality score. High-quality frontal poses produce focused attention on facial features, while degraded conditions show diffuse patterns, consistent across architectures and losses.

\subsection*{Cross-Dataset Analysis}

Figures \ref{fig:datasets_vits_arcface} and \ref{fig:datasets_vitb_adaface} analyze eight benchmarks with 40 images each (8 datasets × 5 images). Figure \ref{fig:datasets_vits_arcface} uses ViT-S/ArcFace; Figure \ref{fig:datasets_vitb_adaface} uses ViT-B/AdaFace. Results demonstrate generalization across diverse distributions, interpretable spatial patterns, and consistency across architectures and training objectives, validating that attention patterns inherently encode quality information.

\subsection*{Error-versus-Discard Curves}

Figure \ref{fig:fnmr4} presents EDC curves for FNMR@FMR=$1e-4$ comparing \ourmethod against SOTA methods across eight benchmarks (LFW, AgeDB-30, CFP-FP, CALFW, Adience, CPLFW, XQLFW, IJB-C) using four FR models (ArcFace, ElasticFace, MagFace, CurricularFace), demonstrating competitive performance with progressive error reduction as low-quality samples are discarded.

\subsection*{Quality Score Distributions}

Figure \ref{fig:quality_dist} shows quality score distributions across evaluation benchmarks, comparing \ourmethod with SOTA methods. All scores are normalized to [0, 1] range, revealing how different methods distribute quality predictions across datasets.

\end{document}